# OpenTuringBench: An Open-Model-based Benchmark and Framework for Machine-Generated Text Detection and Attribution

Lucio La Cava, Andrea Tagarelli
DIMES Dept., University of Calabria
v. P. Bucci 44Z, 87036 Rende, CS, Italy
{lucio.lacava,tagarelli}@dimes.unical.it

## Abstract

*Open* Large Language Models (OLLMs) are increasingly leveraged in generative AI applications, posing new challenges for detecting their outputs. We propose OpenTuringBench, a new benchmark based on OLLMs, designed to train and evaluate machine-generated text detectors on the Turing Test and Authorship Attribution problems. OpenTuringBench focuses on a representative set of OLLMs, and features a number of challenging evaluation tasks, including human/machine-manipulated texts, out-of-domain texts, and texts from previously unseen models. We also provide OTBDetector, a contrastive learning framework to detect and attribute OLLM-based machine-generated texts. Results highlight the relevance and varying degrees of difficulty of the OpenTuringBench tasks, with our detector achieving remarkable capabilities across the various tasks and outperforming most existing detectors.

Resources are available on the OpenTuringBench HuggingFace repository.

## 1 Introduction

The widespread presence of machine-generated text (MGT) across the Internet and various communication channels has nowadays reached unprecedented levels, driven by the rapid advancements in generative AI tools based on large language models. Today, machines can mimic humans (La Cava and Tagarelli, 2025), generating text with impressive realism and contextual relevance and flooding human communications while remaining undetected (Jakesch et al., 2023).

The abundance of MGT impacts the reliability, credibility, and trustworthiness of information sources, thus exposing their users to several problems, such as misinformation and fake news (Yang and Menczer, 2024; Chen and Shu, 2024), plagiarism and content authenticity, loss of content originality, and contamination of training data for future AI models (Shumailov et al., 2023).

To address the above challenges, *MGT detectors* play a key-enabling role. However, it is essential that the development of such detectors is coupled with benchmarks to assess their effectiveness and timeliness, thus raising the standards to match the fast-evolving capabilities of modern models of generating human-like text (Wu et al., 2025). These have become increasingly accessible yet diversified due to the emergence of ***open* large language models** (OLLMs). With the term *open*, here we refer to models distributed under a permissive license, granting free use and/or unrestricted access to the models' weights and documentation, possibly at different degrees and under different modalities.

**Why focus on OLLMs.** One fundamental reason lies in the ethical responsibility of every scholar to uphold open research principles, which would ensure that scientific advancements remain a shared, collective good rather than proprietary assets.

Beyond that, it should be acknowledged that OLLMs are nowadays a competitive alternative to commercially-licensed models in many applications. OLLMs are not only being released with unprecedented volume—as of early 2025, the HuggingFace Hub hosts more than 170K text-generation "open" models—but the ease of *locally running and hosting* these models without the requirement of sending data to the servers of the model's owner, thus ensuring full data privacy, enhanced security, and greater control over sensitive information, aligning with ethical and regulatory standards for responsible AI usage.

All of the above aspects are crucial to foster the creation and spread of unlimited MGT contents. Interestingly, a raising trend is the exploitation of OLLMs as generators of synthetic data for the training of newer, larger (open) models; for example, NVIDIA utilizes several OLLMs to generate training data for its Nemotron family (Adler et al., 2024; Blakeman et al., 2025), while DeepSeek resorts to their R1 model to distill knowledge into the latest DeepSeek-v3 (Liu et al., 2024). From this perspective, compared to closed models, OLLMs make MGT detection—particularly authorship attribution—more challenging due to the increased diversity in their architectures and training data. This difficulty is underscored by recent work (Dugan et al., 2024), which shows that outputs from LLM are very difficult to detect, especially in the case of non-chat OLLMs.

**Contributions.** Despite their widespread use, however, OLLMs remain underexplored in the context of MGT detection, particularly when it comes to the most recent and advanced models. This gap can largely be attributed to the lack of suitable training data and the scarcity of robust benchmarks specifically designed for their evaluation. Therefore, there is an urgent need to advance MGT detection tools and benchmarks for OLLMs, which is the main focus of this work. Our main contributions can be summarized as follows:

• We propose OpenTuringBench, a novel and large-scale ($>$ 500K texts) benchmark specifically designed to train and evaluate MGT detectors on *open* large language models.

• Upon the fundamental MGT-detection problems, i.e., Turing Test and Authorship Attribution, OpenTuringBench involves a set of 7 evaluation tasks that reflect scenarios of increasing levels of difficulty, such as the detection and attribution of human/machine-manipulated text, out-of-domain text, and texts generated from unseen models.

• We develop OTBDetector, a contrastive-learning-based framework trained on OpenTuringBench to detect and attribute OLLM-based MGT.

• Our experimental results on MGT detection and attribution show the relevance and varying degrees of difficulty of the OpenTuringBench evaluation tasks, and highlight the significance of OTBDetector across the various tasks, outperforming all of the 9 competing detectors.

**Comparison with Existing Benchmarks.** Our OpenTuringBench features a number of key novelties w.r.t. existing benchmarks (cf. Sect. 7):

• *Broader coverage of OLLMs*: In building the OpenTuringBench, we generate both training and evaluation texts using a diverse set of recently released models (from 2024 or late 2023), spanning multiple model families and parameter scales. By contrast, prior benchmarks such as (Uchendu et al., 2021) rely on quite outdated models (with the most recent dating back to 2020), while others like (Wu et al., 2024), include only a limited selection of OLLMs, offering a much narrower view of the current LLM landscape.

• *Challenging tasks*: OpenTuringBench poses particular emphasis on authorship attribution, unlike most existing benchmarks. Particularly, it provides a unique framework since it addresses both MGT detection and attribution in challenging evaluation scenarios, including *mixed human-machine*, *out-of-domain*, and *previously unseen models'* texts. Existing works either focus on the conventional attribution task using in-domain only data (Uchendu et al., 2021), or consider some challenging tasks in a detection-only benchmark setting (Dugan et al., 2024), or focus on multilingual MGT detection (Macko et al., 2023).

• *Larger size*: OpenTuringBench is significantly larger than most existing benchmarks (e.g., 7x w.r.t. (Macko et al., 2023)), with more samples per generator (e.g., more than 6x w.r.t. (Uchendu et al., 2021)).

• *Baseline detector*: Unlike traditional benchmarks, we also release a dedicated baseline detector designed to evaluate current detection systems trained on OpenTuringBench and future ones, thus providing a standardized point of comparison.

## 2 OpenTuringBench

### 2.1 Data Creation

**Domain choice.** Choosing the discourse domain for building an MGT detection benchmark is crucial, having several implications on the quality of the benchmark data and on the significance of the detection tasks. We believe that choosing the domain of *news articles* is a strategic decision due to several important factors, including reliability and factuality of the generated contents, diversity of writing styles and topics, challenges of coherence and realism, ethical implications about authorship, transparency, and accountability.

**Data source.** Within this view, we resorted to the *News Category* dataset,[1] one of the largest publicly available datasets of news headlines and articles from *HuffPost*, spanning between 2012 and 2022. This choice is mainly motivated by the following reasons: *(i)* broad spectrum of knowledge, as the dataset covers 42 different subjects, from 'politics' to 'entertainment'; *(ii)* temporal cut-off ensuring that the news data are human-generated, since advanced generative AI models and platforms (e.g., ChatGPT) were not yet released at that time.

---
[1] https://www.kaggle.com/datasets/rmisra/news-category-dataset

For each of the 42 subject categories of the *News Category* dataset, we selected 1000 news headlines (or the maximum available, when fewer than 1K were available), obtaining a total of 41,426 human-authored news headlines and associated articles, covering 11,615 different journalists. The word length of headlines ranges from 1 to 29, with mean 9.63 and standard deviation 3.01, respectively.

**Generation models.** Our proposed OpenTuring-Bench involves a representative body of the OLLM landscape, varying by sizes and architectures, for which we accessed their publicly available implementations on the *HuggingFace Model Hub* as of late 2024, namely: **Llama3**, **Gemma2**, **Qwen**, **Mistral**, **Phi3**, **NeuralChat**, and **SOLAR**. Full details on models and settings can be found in ***Appendix A***.

**Machine-generated data.** We prompted each LLM to generate a news article given a headline. Details on prompts are reported in ***Appendix B***. We also constrained the LLMs to align with the news' date specified in the prompt, to improve factual accuracy in the generated content. Moreover, we noticed no generation refusal behaviors exhibited by the OLLMs under study, which would indicate no presence of harmful topics in the MGTs.

We eventually performed a data cleaning step to remove any degenerated generation. This resulted in 289,982 machine-generated news articles, for a total of 331,408 including the 41,426 human-written ones. This collection was split into *train-val-test* sets following an 80-10-10 ratio for the subsequent training and evaluation tasks.

**OpenTuringBench comprises a total of 543,091 texts**, including the test sets corresponding to the evaluation tasks, as summarized in Table 1.

### 2.2 Data Exploration

We examined properties of our benchmark texts using *internal* and *external* statistical criteria: the former refer to measurements of the characteristics of the human- and machine-generated texts, while external criteria compare the machine-generated texts with the human-written texts corresponding to the same headlines. Here we overview such criteria; please refer to ***Appendix C*** for further details.

For the internal criteria, we focus on style and quality aspects of a MGT. We compute (i) **baseline counts** (no. of syllables, words and sentences), (ii) **compression ratio** (i.e., original size divided by gzip compressed size), (iii) **readability**, which indicates how difficult a passage in English is to understand in function of the counts of words, syllables, and sentences in the text,[2] and (iv) **part-of-speech distribution heterogeneity**. For readability, we resort to two methods: the *Flesch Reading Ease* score, which ranges within (-∞, 121.22], where higher values correspond to better readability; and the *Readability Consensus* score, which estimates the school grade level required to understand an input text, where higher values correspond to higher grades, i.e., lower readability. Part-of-speech distribution heterogeneity is assessed through **POS-entropy**, which is originally introduced in this work as an entropy measurement of the distribution of part-of-speech items associated with the words in a generated text. We also introduce a weighted variant, named *positional POS-entropy*, such that weights are assigned to the positions of POS items using an exponential decay function.

Concerning external criteria, we account for syntactic and semantic similarity analysis based on (i) **Edit distance**, (ii) **Compression ratio** of the concatenation of the human-written and machine-generated text, (iii) $n$-**gram diversity** (Meister et al., 2023), (iv) **self-repetition**,[3] and (v) **homogenization** scores (Padmakumar and He, 2024) using BLEU, ROUGE-L and BERTscore methods.

**Summary of results.** We analyzed our benchmark data according to the above criteria; results are reported in ***Appendix C***—Tables 6–7 for the train set, and Table 8 for the test set. Compared to the human-written texts, MGTs tend to be shorter, less compressible, less readable, and with slightly less heterogeneity of POS patterns. In addition, it stands out that ROUGE-L and BLEU scores are extremely low, indicating significant differences in wording and phrasing from the human-written articles, while the moderate BERTscore suggests that the machine-generated articles might capture some high-level concepts of the human-written text but, as expected, likely misses important nuances like the actual facts of the original news.

### 2.3 Training Tasks

We are given a set $\mathcal{X} = \{X_i\}_{i=1}^N$ of texts (e.g., news articles) generated by humans and machines.

---

[2] https://pypi.org/project/textstat/
[3] https://pypi.org/project/diversity/

| Goal | Task | Test data | # Models (Tested) | # Classes TT | # Classes AA |
|------|------|-----------|-------------------|----|----|
| | | Train data 264,321 | Validation data 33,040 | | |
| ID | E0 | 33,051 | 7 + 1 | 2 | 8 |
| ID-V | E1 | 66,102 | 7 + 1 | 2 | 8 |
| | E2 | 33,051 | 7 + 1 | 2 | 8 |
| | E3 | 33,042 | 7 + 1 | 2 | 8 |
| | E4 | 65,718 | 7 + 1 | 2 | 8 |
| OOD | E5 | 6,573 | 7 | 1 | 7 |
| | E6 | 8,193 | 1 + 1 | 2 | 2 |

Table 1: Summary of OpenTuringBench data splits and tasks. ID, resp. ID-V, indicates in-domain and in-domain variations, respectively; OOD indicates Out-of-Distribution tasks. "+1" in # Models indicates the availability of human-generated data.

If we denote with $y_h$ the 'HUMAN' class label and with $\mathcal{Y}_m = \{y_j\}_{j=1}^{M}$ the set of 'MACHINE' class labels, the training task is to learn a classification model, supervisedly trained on the instances of $\mathcal{X}$ and their class labels, capable of predicting the class for any given unseen text sample.

This task can be formulated as two distinct problems: a binary classification problem, known as *Turing Test* (**TT**), i.e., to distinguish between human- and machine-generated text (without requiring any identification of the specific generator), by learning a mapping function $f : \mathcal{X} \mapsto \{0,1\}$, where 0 corresponds to $y_h$ and 1 to *any* of the labels in $\mathcal{Y}_m$; a multi-class classification problem, known as *Authorship Attribution* (**AA**), i.e., to recognize the author of a given text choosing between a human ($y_h$) or $M$ machine-generators, by learning a mapping function $f : \mathcal{X} \mapsto \mathcal{Y} = \{y_h\} \cup \mathcal{Y}_m$.

## 2.4 Evaluation Tasks

To assess the performance of detectors being trained on our OpenTuringBench, for both TT and AA problems, we introduce the following evaluation goals: *(i)* **in-domain** (ID) tasks, which involve evaluating on test news articles from the same data source of OpenTuringBench or newly generated texts under different settings/prompts of the models, and *(ii)* **out-of-distribution** (OOD) tasks, i.e., testing on texts from a *different domain* than news articles or generated by *previously unseen models*.[4] Table 1 provides an overview of these tasks, which we elaborate on next.

---
[4]All evaluation tasks are considered under both binary (i.e., TT) and multi-class (i.e., AA) classification settings.

**E0: Turing Test and Authorship Attribution.**
These correspond to the evaluation on the test set of news articles of OpenTuringBench, according to the primary goals of assessing the ability of AI detectors to distinguish MGT from human-authored content (i.e., TT) and to recognize the authorship of the content generator (i.e., AA).

**E1: Impact of Temperature.** As is well-known, higher temperature leads to increasing creativity and diversity in the generated output. To assess the impact of temperature on the models' performance, we produced alternative sets of news corresponding to our test headlines by setting the sampling temperature to 0.7 and 1.0, respectively. These lead to higher randomness in generations, thus potentially challenging detectability, while ensuring a proper balance between coherence and creativity in the generated news articles.

**E2: Impact of Model Size.** This task evaluates the impact of model size on the detection and attribution of MGT. Indeed, larger models, with their greater number of parameters, might generate different content due to broader knowledge. We utilized the $\sim 70B$ implementations of *Llama 3.1* and *Qwen 2.5* to generate a new test set aimed at assessing whether detectors trained on their smaller versions (i.e., 8B and 7B, respectively) can also distinguish these models in a larger size context.

**E3: Impact of Text Rewriting.** This evaluation task explores how text manipulation impacts the detection and attribution performance, with each model rewriting its own previously generated texts. This might lead the model to introduce/remove subtle patterns or noise into/from the previous generation, thus getting confused detection systems that are trained to recognize specific patterns.

**E4: Human-Machine Mixing.** This aims to investigate how combinations of human-generated and MGT impact the detection and attribution performance (Tripto et al., 2024). For each model in Table 4, we propose the following scenarios, each corresponding to a new distinct test set:

• **Human-Content Revision:** Human-written content is revised by each model. This introduces challenges as *(i)* machine-generated revisions may blend with human text, thus remaining undetectable and, conversely, *(ii)* the human-generated text can retain most of its patterns, thus complicating detection of machine edits.

- **Human-Content Continuation:** Models are asked to continue human-authored content, which is expected to further complicating the challenges introduced in human-content revision.

**E5: Out-of-Domain Text.** This task examines how effectively detection and attribution capabilities generalize to texts generated in a domain that differs from the training one (i.e., news). We choose the *essay* domain, which presents different linguistic patterns or jargon, while still requiring high coherence and reliability of the information generated. To this aim, each of the OLLMs was asked to generate about 1000 essays, using the prompts provided in (Verma et al., 2024).

**E6: Previously-unseen Models.** This task assesses the robustness of detectors against MGTs produced by a model not involved in OpenTuringBench. This task is both crucial and challenging, as it measures the ability of a detector to generalize to new text-generation models, regardless of being open or closed. For this task, we choose *Yi-1.5-9B-Chat* as the "previously unseen" model. Indeed, built upon the *Llama* architecture, we expect that *Llama* (seen during training) would be detected as the closest generation model.

## 3 The OTB Benchmark Detector

Here we present our developed MGT-detector, dubbed OTBDetector, specifically trained on OpenTuringBench.

**Learning framework.** At its core, the OTBDetector architecture follows the promising approach proposed in (La Cava et al., 2024), which performs *similarity learning* to learn a latent similarity space where deeply contextualized representations of texts with a shared class label (i.e., authorship category) are kept close together, whereas those having different labels are pushed far apart. Therefore, human-written texts are expected to be closer to each other than machine-generated ones and, similarly, MGTs from a specific model are expected to stay closer to each other than those generated by other models.

OTBDetector consists of three key components: *(i)* a Pre-Trained Language Model (PLM) to encode the input text data; *(ii)* a *triplet network* architecture, which exploits *contrastive learning* (Bromley et al., 1993) to induce the similarity space between embeddings according to a contrastive loss function; and *(iii)* a *nearest centroid classifier* module to assign query text to the closest authorship category in the learned similarity space.

It should be noted that, differently from (La Cava et al., 2024), we used the *Longformer* model[5] as encoder. This is motivated since the MGTs in OpenTuringBench are usually longer than 1k tokens, which raises a need for PLMs capable of processing longer sequences than the usual BERT's 512-token limit. Longformer efficiently addresses this need due to its strong performance on long-document tasks, including classification (Beltagy et al., 2020).

**Training.** We start by constructing triplets of textual data objects to be fed into the triplet network, consisting of an *anchor* $X^{(a)}$, a *positive sample* $X^{(p)}$, and a *negative sample* $X^{(n)}$, where the positive sample shares the same category of the anchor (i.e., $y^{(p)} = y^{(a)}$), whereas the negative sample belongs to a different category (i.e., $y^{(n)} \neq y^{(a)}$).

Each text $X_i \in \mathcal{X}$ is fed through the *tokenization* process of the PLM associated with OTBDetector to obtain the corresponding token sequence $T_i = [\tau_{i,1}, \ldots, \tau_{i,|T_i|}]$, which is lately mapped into a dense, relatively low dimensional, space of size $f$. Finally, the resulting *token embeddings* of $T_i$, i.e., $PLM(T_i) \in \mathbf{R}^{f \times |T_i|}$, are converted into a single representation, or *sentence embeddings*, using an *average pooling function* that produces an embedding vector $\mathbf{h}_i$ of size $f$:

$$\mathbf{h}_i = pooling(\mathsf{PLM}(T_i)) \in \mathbf{R}^f. \quad (1)$$

The embeddings $\mathbf{h}^{(a)}, \mathbf{h}^{(p)}, \mathbf{h}^{(n)}$ of the anchor, positive and negative objects, respectively, computed by Eq. 1, are fed into the triplet network, which is responsible for optimizing the *triplet loss*, i.e., minimizing the distance between an anchor and a positive, both having the same category and maximizing the distance between the anchor and a negative of a different category:

$$\mathcal{L} = \sum_{\langle X^{(a)}, X^{(p)}, X^{(n)} \rangle} \max(d(\mathbf{h}^{(a)}, \mathbf{h}^{(p)}) - d(\mathbf{h}^{(a)}, \mathbf{h}^{(n)}) + \lambda, 0) \quad (2)$$

where $d(\cdot, \cdot)$ is a distance function and $\lambda \in \mathbf{R}^+$ is a margin between positive and negative pairs.

**Inference.** OTBDetector first pre-computes offline the centroids for each category $y_j \in \mathcal{Y}$ seen during training on $\mathcal{X}$. These are defined as $\mathbf{y}_j = (1/|\mathcal{X}_j|) \sum_{X_i \in \mathcal{X}_j} \mathbf{h}_i$, where $\mathcal{X}_j$ denotes the subset of $\mathcal{X}$ containing data objects of category $y_j$, and $\mathbf{h}_i$ the embedding of the data object $X_i$.

---
[5] https://huggingface.co/allenai/longformer-base-4096

To label an unseen data object $X$, OTBDetector computes its embedding $\mathbf{h}$ according to Eq. 1, and compares it to the pre-computed centroids assigning $X$ with the label $y_j^*$ corresponding to the nearest centroid, such that $k^* = \arg\min_{k=1..M} d(\mathbf{h}, \mathbf{y}_k)$, where $d(\cdot)$ represents a distance metric, in our case the cosine similarity.

## 4 Experimental Setup

**Competing Methods.** To assess the challenges introduced by OpenTuringBench and validate the performance of the proposed OTBDetector, we resorted to the most widely adopted detection methods, which also include those previously collected and provided by MGTBench (He et al., 2024):[6]

• *Metric-based* detection methods, i.e., **Log-Likelihood** (Solaiman et al., 2019), **Rank**, **Entropy**, **GLTR** (Gehrmann et al., 2019), **Log-Rank** (Mitchell et al., 2023), **LRR** (Su et al., 2023), and **Fast-DetectGPT** (Bao et al., 2024).

• *Model-based* detection methods, i.e., **OpenAI Detector** (Solaiman et al., 2019), **ChatGPT Detector** (Guo et al., 2023), **LM Detector** (He et al., 2024), and **DeTeCtive** (Guo et al., 2024).

Consistently with MGTBench, metric-based models exploit the GPT2-medium model, with a logistic regression module on top of it for the subsequent classification tasks. Furthermore, it is worth noting that, to ensure a fair evaluation, we fine-tuned the model-based methods for 10 epochs on our OpenTuringBench train set. This step is necessary, particularly under the Authorship Attribution scenario, to adapt the models to the class structure of our data, which differs from their original training data. Details on the above competing detectors can be found in ***Appendix F***.

**Evaluation Metrics.** We assessed the performance of OTBDetector and competing methods through standard metrics derived from the confusion matrices obtained in the various TT and AA tasks (cf. Sect. 2.3). These include *precision* ($P$), *recall* ($R$), and $F_1$-*score* ($F_1$), where outcomes 'MACHINE' are regarded as instances of the *positive* class. The scores were computed using a *weighted-average* approach, which is commonly used to account for variations in class sizes and to align with the default settings in MGTBench.

---
[6]We referred to the public implementations available at https://github.com/TrustAIRLab/MGTBench/

Note that all detectors were always trained only on the train split of OpenTuringBench, and evaluated against the test-sets corresponding to the various OpenTuringBench tasks (cf. Table 1).

## 5 Results

We organize the presentation of results achieved by OTBDetector and competing methods into two parts: a summary of the methods' performance for the TT problem, and a detailed presentation on the more challenging AA problem under *(i)* In-Domain Tasks, *(ii)* variations of In-Domain Tasks, and *(iii)* Out-of-Distribution Tasks.

### 5.1 Turing Test (summary)

Due to page limitations, here we present a summary of the results on TT rather than a detailed discussion—which is available in ***Appendix G*** (cf. upper subtables of Tables 9-10)—as TT turned out to be apparently easy, leading all evaluated methods to achieve remarkably high performance. In fact, we observe $F_1$ scores consistently exceeding 0.9, especially in the case of LM-D and OTBDetector. This actually comes with no surprise, as it is in line with our data exploration (Sect. 2.2) that revealed substantial divergences in linguistic patterns between humans and machines, enabling both *metric*-based and *model*-based approaches to leverage these differences effectively.

### 5.2 Authorship Attribution

**In-Domain Benchmark Tasks (E0).** Compared to TT, AA represents a more challenging scenario. Indeed, metric-based approaches struggle to accurately attribute text to the correct model, with 0.343 $F_1$ in the best case. This difficulty arises because, while there are evident divergences in patterns between humans and machines (cf. Table 6), such distinctions are far less pronounced among different machines, as further detailed in ***Appendix C***. By contrast, model-based approaches consistently achieve strong performance, with OTBDetector emerging as the best one with an $F_1$ score of 0.996.

**In-Domain Variations of Benchmark Tasks (E1 to E4).** We first assessed the impact of raising the models' temperature during generation (**E1**). As shown in the second and third left-most columns of Table 2, increasing the temperature to 0.7 does not affect performance of most models, with LM-D and OTBDetector maintaining their effectiveness. However, detectors like OAI-D and GPT-D exhibit notable sensitivity to the increased randomness in generation, with reductions in $F_1$ of 27% and 19%, respectively, likely due to less robustness to pattern variations in the generated texts. This becomes more evident with temperature 1.0, with all detectors experiencing a further decrease in performance. Nonetheless, OTBDetector still achieves the best performance with 0.838 $F_1$.

| Test Task | Default | | | Higher Temp (0.7) | | | Higher Temp (1.0) | | | Larger Size | | | Self-Rewriting | | | Human Revision | | | Human Contin. | | |
|---|---|---|---|---|---|---|---|---|---|---|---|---|---|---|---|---|---|---|---|---|---|
| Detector | P | R | $F_1$ | P | R | $F_1$ | P | R | $F_1$ | P | R | $F_1$ | P | R | $F_1$ | P | R | $F_1$ | P | R | $F_1$ |
| Log-L | 0.305 | 0.324 | 0.299 | 0.230 | 0.253 | 0.210 | 0.161 | 0.154 | 0.078 | 0.292 | 0.306 | 0.286 | 0.249 | 0.281 | 0.241 | 0.172 | 0.137 | 0.064 | 0.158 | 0.147 | 0.075 |
| Rank | 0.211 | 0.235 | 0.190 | 0.203 | 0.236 | 0.191 | 0.168 | 0.147 | 0.106 | 0.202 | 0.226 | 0.184 | 0.197 | 0.229 | 0.180 | 0.133 | 0.140 | 0.096 | 0.139 | 0.152 | 0.106 |
| Log-R | 0.324 | 0.338 | 0.320 | 0.235 | 0.250 | 0.207 | 0.182 | 0.154 | 0.075 | 0.304 | 0.314 | 0.301 | 0.262 | 0.293 | 0.259 | 0.188 | 0.135 | 0.059 | 0.147 | 0.149 | 0.075 |
| Entropy | 0.162 | 0.188 | 0.141 | 0.153 | 0.173 | 0.129 | 0.131 | 0.129 | 0.093 | 0.161 | 0.187 | 0.140 | 0.156 | 0.185 | 0.136 | 0.127 | 0.134 | 0.093 | 0.127 | 0.125 | 0.087 |
| GLTR | 0.345 | 0.345 | 0.343 | 0.246 | 0.243 | 0.209 | 0.165 | 0.145 | 0.073 | 0.313 | 0.324 | 0.315 | 0.294 | 0.305 | 0.283 | 0.196 | 0.134 | 0.061 | 0.184 | 0.152 | 0.085 |
| LRR | 0.309 | 0.330 | 0.310 | 0.239 | 0.241 | 0.193 | 0.176 | 0.120 | 0.048 | 0.283 | 0.299 | 0.283 | 0.267 | 0.292 | 0.259 | 0.152 | 0.121 | 0.043 | 0.156 | 0.154 | 0.076 |
| OAI-D | <u>0.990</u> | <u>0.990</u> | <u>0.990</u> | <u>0.975</u> | <u>0.975</u> | <u>0.975</u> | 0.837 | 0.800 | 0.783 | 0.986 | <u>0.985</u> | <u>0.985</u> | 0.894 | 0.855 | 0.851 | <u>0.500</u> | **0.331** | **0.280** | 0.466 | <u>0.187</u> | <u>0.110</u> |
| GPT-D | 0.986 | 0.986 | 0.986 | 0.965 | 0.964 | 0.964 | 0.835 | 0.806 | 0.791 | 0.978 | 0.978 | 0.978 | 0.885 | 0.806 | 0.800 | 0.404 | 0.297 | 0.240 | 0.388 | 0.180 | 0.093 |
| LM-D | <u>0.990</u> | <u>0.990</u> | <u>0.990</u> | 0.969 | 0.969 | 0.969 | 0.859 | 0.840 | 0.830 | 0.978 | 0.977 | 0.977 | <u>0.895</u> | 0.849 | 0.849 | 0.434 | 0.187 | 0.119 | **0.573** | 0.145 | 0.068 |
| DeTeCtive | **0.997** | **0.997** | **0.997** | **0.983** | **0.983** | **0.983** | **0.903** | **0.890** | **0.886** | <u>0.990</u> | 0.742 | 0.742 | **0.914** | <u>0.871</u> | <u>0.864</u> | **0.542** | 0.291 | 0.245 | 0.469 | 0.174 | 0.096 |
| Ours | **0.997** | **0.997** | **0.997** | **0.983** | **0.983** | **0.983** | <u>0.874</u> | <u>0.845</u> | <u>0.838</u> | **0.995** | **0.995** | **0.995** | 0.889 | **0.877** | **0.871** | 0.483 | <u>0.315</u> | <u>0.286</u> | <u>0.518</u> | **0.238** | **0.148** |

Table 2: ID and ID-V tasks (**E0**-**E4**) for Authorship Attribution. Best scores are in bold, second-best underlined.

When considering models of larger size (**E2**), the detectors exhibit comparable trends in relative performance to those observed for the default AA scenario (cf. central column in Table 2). With the exception of OTBDetector (which is again the best-performer, with 0.995 $F_1$) and Entropy, all the other models achieve decreased performance, although less severely compared to the variation in temperature. This would indicate that even with more parameters and greater knowledge, models still rely on similar patterns while generating texts.

The self-rewriting task (**E3**) further challenges detectors by stressing their generalization capabilities as generation patterns are perturbed due to rephrasing. This task indeed results in a drastic decrease in performances, as shown in the third right-most column of Table 2. This is particularly evident for model-based approaches, which rely on semantic patterns that are now altered by rephrasing. Nonetheless, OTBDetector continues to outperform others (0.871 $F_1$).

Much more challenging is mixing human-generated and machine-generated content (**E4**), which significantly disrupts detectors' performance, as shown in the last two columns of Table 2. OTB-Detector is the best performer and model-based detectors maintain relatively higher robustness compared to metric-based detectors, for the *human revision subtask*, although with $F_1$ never exceeding 0.3. The difference between the two types of detectors then becomes nearly indistinguishable in the *human continuation subtask*, where the maximum $F_1$ (achieved by our detector) is only 0.15, highlighting the nature of the task as one of extreme classification for all detectors involved.

| Test Task | Out-of-Domain Text | | | Unseen Model | | |
|---|---|---|---|---|---|---|
| Detector | P | R | $F_1$ | P | R | $F_1$ |
| Log-L | 0.283 | 0.295 | 0.256 | 0.990 | 0.515 | 0.607 |
| Rank | 0.170 | 0.169 | 0.120 | 0.953 | **0.625** | **0.755** |
| Log-R | 0.294 | 0.311 | 0.280 | 0.995 | 0.516 | 0.603 |
| Entropy | 0.169 | 0.155 | 0.113 | 0.533 | 0.383 | 0.434 |
| GLTR | 0.251 | 0.260 | 0.248 | 0.992 | 0.492 | 0.571 |
| LRR | 0.230 | 0.235 | 0.223 | <u>0.998</u> | 0.515 | 0.597 |
| OAI-D | 0.569 | 0.439 | 0.429 | 0.996 | 0.567 | 0.622 |
| GPT-D | 0.497 | 0.400 | 0.392 | 0.991 | 0.507 | 0.521 |
| LM-D | 0.589 | 0.343 | 0.298 | <u>0.998</u> | 0.531 | 0.565 |
| DeTeCtive | <u>0.615</u> | <u>0.454</u> | <u>0.456</u> | **0.999** | 0.494 | 0.494 |
| Ours | **0.667** | **0.548** | **0.510** | **0.999** | <u>0.595</u> | <u>0.663</u> |

Table 3: OOD tasks (**E5**-**E6**) for Authorship Attribution. Best scores are in bold, second-best underlined.

**Results on Out-of-Distribution Benchmark Tasks (E5-E6).** Finally, we evaluated how detectors trained on OpenTuringBench generalize their capabilities to unseen domain or machine-generators, as reported in Table 3.

The out-of-domain (i.e., essay) scenario (**E5**), significantly challenges generalization capabilities, causing a sharp reduction in performance, however less severe than in the previously discussed **E4** tasks. OTBDetector remains the top performer, with 0.51 $F_1$, surpassing OAI-D by +18%, and doubling the scores of most other detectors.

When examining generalization to machine-generators not involved during training (**E6**), the scenario appears to be generally less challenging for the detectors. With the exception of Entropy, all detectors exhibit precision close to 1, however with $F_1$ around 0.6. Our OTBDetector achieves 0.66 $F_1$ and is second only to Rank, which has better recall; this might be due to the two-class setting of this AA task (cf. Table 1), which leads Rank to make fewer false negative errors, i.e., detect 'HUMAN' instead of 'MACHINE'.

## 6 Discussion

**Benchmark Tasks Summary.** OpenTuringBench enables effective training of detectors capable of identifying and attributing OLLM-generated text, as demonstrated by the strong performance achieved by the experimented detectors across the conventional tasks of TT and AA. However, the novel tasks we introduced in OpenTuringBench pose significant challenges to the detectors, particularly unveiling their limitations in generalizing to mixed human-machine texts and to out-of-domain texts. We believe this highlights the urgent need for next-generation MGT detection approaches.

**Detectors Summary.** Benchmarking multiple detectors on OpenTuringBench also provided valuable insights into their strengths and weaknesses. Metric-based detectors have shown to behave well in capturing linguistic pattern variations that are relatively robust to more challenging TT tasks due to their ability to identify clear differences among human-generated and machine-generated content. Conversely, model-based detectors are superior for AA tasks, as the lack of significant linguistic differences among machine generators (cf. *Appendix C*) makes semantic patterns more effective for this purpose. Notably, our proposed OTBDetector emerges as the most effective attribution method, consistently achieving superior performances across different evaluations, demonstrating stronger resilience to more challenging tasks.

## 7 Related Work

The LLMs' ability to generate coherent, creative, and contextually relevant text has contributed to determine great interest in MGT detection (Jawahar et al., 2020; Wu et al., 2023).

Watermarking approaches have attracted some attention due to their capability to embed latent signals into MGT that remain hidden to humans, yet still detectable by machines (Kirchenbauer et al., 2023; Yoo et al., 2023). Statistical methods provide multifaceted approaches to determine whether a text has been machine-generated, including rank-related scores (Mitchell et al., 2023; Su et al., 2023), entropy (Gehrmann et al., 2019), discourse motifs (Kim et al., 2024), along with other statistical approaches (Tulchinskii et al., 2023; Wang et al., 2023; Venkatraman et al., 2024; Bao et al., 2024). Deep learning frameworks have also proven promising for detecting MGT (Ippolito et al., 2020; Verma et al., 2024; Bhattacharjee and Liu, 2024; Uchendu et al., 2024). In this context, a particularly promising approach to MGT detection is based on contrastive learning (La Cava et al., 2024), through domain adaptation (Bhattacharjee et al., 2023) or adversarial training (Bhattacharjee et al., 2024).

The above methods have traditionally considered MGT produced by closed or commercially-licensed models. This focus has also been reflected in the development of benchmarks and evaluation frameworks. *TuringBench* (Uchendu et al., 2021) is one of the earliest efforts for supporting MGT detection and attribution. The *Human ChatGPT Comparison Corpus* (HC3) (Guo et al., 2023) offers a collection of 40K questions and answers, enabling the analysis of ChatGPT and humans' linguistic aspects. *MULTITuDE* (Macko et al., 2023) and *MultiSocial* (Macko et al., 2024) focus on multilingual MGT detection, by providing long and short texts, respectively, generated in different languages by multilingual LLMs, also evaluating detectors in the multilingual context. *DetectRL* (Wu et al., 2024) benchmarks MGT detection under real-world scenarios based on adversarial LLM-generated text. *MGTBench* (He et al., 2024) provides a benchmark tool to assess the performance of detectors for MGT, including their resilience to adversarial attacks, and highlighting the need to develop more robust detection methods.

## 8 Conclusions

We presented OpenTuringBench, a novel benchmark featuring more than 500K texts for the training and evaluation of methods for MGT detection and attribution, based on OLLMs. OpenTuringBench fills a gap in the current literature on MGT detection benchmarks as it combines a broader coverage of OLLMs and challenging evaluation tasks, such as the detection and attribution of human/machine-manipulated text, out-of-domain text, and texts generated from unseen models.

We also presented OTBDetector, a contrastive learning framework to detect and attribute OLLM-based MGT, which has shown to provide better performance than existing detectors across most of the OpenTuringBench evaluation tasks.

Our ongoing work is focusing on *(i)* the extension of OpenTuringBench with new families of OLLMs (e.g., (OLMo et al., 2025)) and *(ii)* the use of OTBDetector for the classification of types of machine interventions over human texts.

## Acknowledgements

AT, resp. LLC, was supported by project "Future Artificial Intelligence Research (FAIR)" spoke 9 (H23C22000860006), resp. project SERICS (PE00000014), both under the MUR National Recovery and Resilience Plan funded by the EU - NextGenerationEU.

## Limitations

**Discourse domains.** As discussed in Section 2, we chose to focus on news articles as our primary discourse domain for a number of reasons. Nevertheless, we recognize the importance of expanding our findings over other discourse domains. Within this view, our ongoing work includes MGT in creative domains by leveraging synthetic personas (Ge et al., 2024), aiming to enhance both diversity and specialization in MGT.

**Language usage.** Our benchmark currently refers only to English texts. Since linguistic features and detectors' performance can vary across languages, following the lead of resources like (Macko et al., 2023), we will look at exploring multilingual models to extend our benchmark and support multilingual tasks.

**Continual learning.** Currently, integrating MGT from newly released LLMs into OTBDetector requires retraining the system. While this process is relatively straightforward, it becomes increasingly inefficient as the number of models grows. To address this, it is worth developing continual contrastive learning frameworks that allow the system to incrementally incorporate MGT from new LLMs without full retraining. This would enhance scalability and adaptability, making OTBDetector more practical in dynamic, evolving scenarios.

## Ethics Statement

**Detectability of MGT Content.** Our findings highlight the difficulty of reliably detecting and attributing MGT in some tasks. This might conceal potentially harmful purposes and potential misuses, with significant impact on individuals, communities, or society. Accordingly, we strongly urge all parties involved to exercise caution and responsibility to ensure the safe and ethical deployment and utilization of these technologies.

**Broader impact.** The main goal of our research is to advance the field of MGT detection by providing a robust benchmark for evaluating detection and attribution capabilities. At the same time, we acknowledge that our work may inadvertently reveal limitations in current detection tools, which could be exploited for malicious purposes. We discard any responsibility for such misuse and stress the importance of responsible and ethical use of these technologies by all actors involved.

## A  MGT Models and Settings

Table 4 summarizes the main details of the MGT models employed in this work. For all models, we ensured the lowest randomness in generation—according to the high requirements of coherence for news contents—by setting the *temperature* to 0.01, and the *top_p* and *top_k* to their default values of 1 and 50, respectively. We used the `vllm` library [7] for inference on a 8x NVIDIA A30 GPU 24GB NVRAM server.

| Model | Abbrev. | Params | Baseline |
|---|---|---|---|
| SOLAR-10.7B-Instruct-v1.0 | SOLAR | 10.7B | Llama-2 |
| gemma-2-9b-it | Gemma | 9.24B | Custom |
| Llama-3.1-8B-Instruct | Llama3-8 | 8.03B | Llama-3 |
| Qwen2.5-7B-Instruct | Qwen-7 | 7.62B | Custom |
| Mistral-7B-Instruct-v0.3 | Mistral | 7.25B | Mistral |
| Neural-chat-7b-v3-3 | NeuralChat | 7.24B | Mistral |
| Phi-3-mini-4k-instruct | Phi3 | 3.82B | Custom |

Table 4: LLMs selected for our study, ordered by decreasing number of parameters, and annotated with their baseline architecture.

## B  Prompts for Machine Generations

> **News Generation Instructions**
>
> **System Message:** *You are a seasoned journalist tasked with crafting an informative news article based on a provided headline, news category, and date. Approach sensitive topics with care, ensuring that all content does not promote misinformation or harmful stereotypes. Do not refuse to write about any topic, even if it is controversial or sensitive, you are required to write about everything as a journalist. Replace placeholders (e.g., names, places, or dates) with contextually appropriate and fictitious details to enhance realism. Your objective is to produce a thorough, well-rounded, and informative article that captivates readers while upholding journalistic integrity, accuracy, and respect for all individuals and communities involved.*
>
> **System Message:** *Please, generate an article starting from the following information:*
> - *News headline:* $<< headline >>$
> - *News category:* $<< category >>$
> - *News date:* $<< date >>$

Figure 1: Prompt instructions for news generation.

> **Self-Rewriting Instructions**
>
> **System Message:** *You are an expert in rewriting text. You are given a text and you are required to rewrite it in a more coherent and readable way. You are allowed to change the structure, style, and tone of the text, as well as the words. You are required to ensure that the original meaning is preserved.*
>
> **Prompt Message:** *Please, rewrite the following text in a more coherent and readable way:* $<< text >>$

Figure 2: Prompt instructions for self-rewriting tasks.

> **Human-Content Revision Instructions**
>
> **System Message:** *You are an expert in revising human-written text. You are given a text and you are required to revise it.*
>
> **Prompt Message:** *Please, revise the following text:* $<< text >>$

Figure 3: Prompt instructions for human-content revision tasks.

> **Human-Content Continuation Instructions**
>
> **System Message:** *You are an expert writer. You are given a text and you are required to write a continuation of it.*
>
> **Prompt Message:** *Please, write a continuation of the following text:* $<< text >>$

Figure 4: Prompt instructions for human-content continuation tasks.

> **Essay Writing Instructions**
>
> **System Message:** *You are an expert in writing essays. You are tasked with crafting an essay. Your objective is to produce a thorough, well-rounded, and informative essay that captivates readers based on the provided instructions.*
>
> **Prompt Message:** $<< essay\ outline >>$

Figure 5: Prompt instructions for essay writing tasks.

## C  Details on Text Statistics

**Flesch Reading Ease.** This is calculated as a linear combination of the average number of words per sentence and the average number of syllables per word. Texts with shorter words and shorter sentences have higher score, i.e., it is easier to un-

---
[7] https://github.com/vllm-project/vllm

| Method | Notation | Type | Params | Library |
|---|---|---|---|---|
| Syllable count | *syC* | int. | – | textstat |
| Lexicon count | *lC* | int. | punct. removal | textstat |
| Sentence count | *sC* | int. | – | textstat |
| Compression ratio | *Cr* | int. | – | diversity |
| Flesch Reading Ease | *FRE* | int. | – | textstat |
| Readability Consensus | *RC* | int. | – | textstat |
| POS entropy | *POS-E* | int. | – | ours |
| positional POS entropy | *pPOS-E* | int. | decay 0.1 | ours |
| Edit distance | *dist* | ext. | – | Levenshtein |
| *Cr* w/ human | *Crh* | int. | – | diversity |
| Homogenization | *hBLEU* | ext. | BLEU | diversity |
| Homogenization | *hROUGE* | ext. | ROUGE-l | diversity |
| Homogenization | *hBERTs* | ext. | BERTScore | diversity |
| $n$-gram Diversity | *n-div* | ext. | $n \in \{1,2,3\}$ | diversity |
| Self-Repetition | *n-SR* | ext. | $n \in \{1,2,3\}$ | diversity |

Table 5: Summary of statistics computed from the news article contents in OpenTuringBench

| Statistic | Human | | | | Machines | | | |
|---|---|---|---|---|---|---|---|---|
| | Mean | Std | Min | Max | Mean | Std | Min | Max |
| *syC* | 751.927 | 496.384 | 0 | 4089 | 722.268 | 125.259 | 179.857 | 2360 |
| *lC* | 511.799 | 333.903 | 0 | 2949 | 445.978 | 73.040 | 130 | 1557.714 |
| *sC* | 27.432 | 18.852 | 1 | 186 | 20.892 | 5.214 | 4.571 | 168.571 |
| *FRE* | 63.770 | 11.791 | -101.290 | 206.840 | 47.319 | 11.351 | -375.491 | 96.151 |
| *RC* | 10.497 | 2.439 | 0 | 35 | 13.415 | 2.506 | 3.143 | 172.286 |
| *POS-E* | 4.480 | 0.101 | 0 | 5.337 | 4.456 | 0.088 | 2.505 | 5.073 |
| *pPOS-E* | 3.725 | 0.187 | 0 | 4.404 | 3.722 | 0.147 | 2.947 | 4.220 |
| *Cr* | 3.054 | 0.472 | 0.056 | 6.636 | 3.491 | 1.919 | 2.428 | 62.252 |
| *dist* | - | - | - | - | 3161.037 | 2083.764 | 993.143 | 52320 |
| 1-*div* | - | - | - | - | 0.527 | 0.051 | 0.156 | 0.761 |
| 2-*div* | - | - | - | - | 1.426 | 0.082 | 0.473 | 1.744 |
| 3-*div* | - | - | - | - | 2.398 | 0.104 | 0.795 | 2.742 |
| 1-*SR* | - | - | - | - | 6.451 | 0.470 | 3.695 | 8.099 |
| 2-*SR* | - | - | - | - | 6.127 | 0.458 | 3.395 | 7.747 |
| 3-*SR* | - | - | - | - | 6.067 | 0.452 | 3.057 | 7.651 |
| *Crh* | - | - | - | - | 2.190 | 0.355 | 1.679 | 16.736 |
| *hROUGE* | - | - | - | - | 0.070 | 0.014 | 0 | 0.142 |
| *hBLEU* | - | - | - | - | 0.009 | 0.010 | 0 | 0.092 |
| *hBERTs* | - | - | - | - | 0.377 | 0.004 | 0.350 | 0.399 |

Table 6: Aggregated statistics from the human- and machine-generated texts in OpenTuringBench train set. Values under column Machines are averages over the various LLMs' aggregated statistics.

derstand it:

$$206.835 - 1.015 \left( \frac{\# \text{ words}}{\# \text{ sentences}} \right) - 84.6 \left( \frac{\# \text{ syllables}}{\# \text{ words}} \right)$$

**$n$-gram diversity.** Given a MGT and its human-written counterpart, $n$-gram diversity (Meister et al., 2023) is computed as follows: (i) concatenate all sentences in the two texts into a single sentence; (ii) tokenize the sentence (standard split by words); (iii) compute all lists of word-based $n$-grams, by varying $n$ from 1 to a specified value; (iii) for each list of size $n = i$, compute the ratio (# unique $n$-grams) / (# $n$-grams).

**Self-repetition** is computed as follows: (i) for each sentence: compute all word-based $n$-grams of a specified size, then for each $n$-gram, compute the number of occurrences in all sentences but the current one, finally sum over the $n$-grams of the current sentence as $ssum$; (ii) compute the logarithm of the $ssum$ over all sentences, and divides the total by the # sentences.

**POS-entropy and positional POS-entropy.** Let **pos** denote a distribution vector of the counts of the part-of-speech (POS) types observed in the input text, i.e., associated with words occurring in the text. In this work, we used the spaCy Python library to extract the POS from a text, based on the English model en_core_web_sm. The *POS-entropy* of **pos** is estimated as follows:

$$- \sum_{h=1..|POST|} freq(pos_h) \log freq(pos_h) \quad (3)$$

where $POST$ denotes the set of POS types observed in the given text, and $freq(pos_h)$ is the relative frequency of the $h$-th POS type.

We also define a variant of POS-entropy which utilizes an exponentially decay weighting function such that the importance of the occurrence of a POS item decreases smoothly with its position. By denoting with $idx(pos_h)$ the list of position indices of the occurrences of the $h$-th POS type in the input text, we define the *positional POS-entropy* as:

$$- \sum_{h=1..|POST|} wfreq(pos_h) \log wfreq(pos_h), \quad (4)$$

where

$$wfreq(pos_h) = \sum_{i \in idx(pos_h)} \hat{w}_i,$$

$\hat{w}_i$ is the normalized weight $w_i = e^{-\alpha i}$, with $\alpha$ as the decay factor, such that smaller values produce smoother weights.

**Homogenization.** In (Padmakumar and He, 2024), the homogenization of a single text is defined as its average pairwise similarity to all other texts written on the same topic. In our setting this reduces to compute the similarity between a MGT and its human-written counterpart, based on ROUGE-L, BLEU, and BERTScore methods. Homogenization metrics range from 0 to 1 with a higher score indicating more similar content.

Tables 7 and 8 provide additional details on aggregated statistics from the train and test sets in OpenTuringBench.

## D  Additional Insights on Differences with Other Benchmarks

Figure 6 compares a machine-generated sample from OpenTuringBench and one from *Turing-*

| Statistic | Human | | | | Gemma | | | | Llama | | | | Mistral | | | |
|---|---|---|---|---|---|---|---|---|---|---|---|---|---|---|---|---|
| | Mean | Std | Min | Max | Mean | Std | Min | Max | Mean | Std | Min | Max | Mean | Std | Min | Max |
| syC | 751.927 | 496.384 | 0 | 4089 | 559.071 | 96.719 | 215 | 1524 | 594.072 | 113.29 | 273 | 2255 | 759.652 | 138.152 | 19 | 2967 |
| lC | 511.799 | 333.903 | 0 | 2949 | 341.708 | 52.759 | 149 | 1006 | 361.702 | 69.225 | 190 | 1769 | 485.002 | 83.74 | 10 | 1737 |
| sC | 27.432 | 18.852 | 1 | 186 | 17.773 | 4.235 | 7 | 69 | 16.409 | 4.72 | 3 | 245 | 22.477 | 5.822 | 1 | 120 |
| FRE | 63.77 | 11.791 | -101.29 | 206.84 | 48.451 | 11.397 | 7.86 | 88.47 | 44.699 | 10.929 | -239.83 | 114.93 | 51.61 | 13.87 | -979.4 | 88.97 |
| RC | 13.044 | 2.191 | 6 | 22 | 14.044 | 2.117 | 2 | 122 | 12.603 | 3.902 | 5 | 405 | 13.797 | 1.766 | 3 | 30 |
| POS-E | 4.48 | 0.101 | 0 | 5.337 | 4.472 | 0.055 | 4.25 | 4.913 | 4.393 | 0.061 | 3.416 | 4.852 | 4.513 | 0.07 | 3.405 | 5.037 |
| pPOS-E | 3.725 | 0.187 | 0 | 4.404 | 3.707 | 0.146 | 2.877 | 4.222 | 3.612 | 0.138 | 3.005 | 4.222 | 3.881 | 0.176 | 2.965 | 4.353 |
| Cr | 3.054 | 0.472 | 0.056 | 6.636 | 3.081 | 0.161 | 2.473 | 5.955 | 3.408 | 0.422 | 2.675 | 33.557 | 3.538 | 0.592 | 1.072 | 49.34 |
| dist | - | - | - | - | 2929.457 | 2193.281 | 759 | 52826 | 2958.623 | 2161.26 | 887 | 52795 | 3178.954 | 1957.849 | 364 | 52024 |
| 1-div | - | - | - | - | 0.563 | 0.061 | 0.246 | 0.772 | 0.527 | 0.05 | 0.103 | 0.704 | 0.51 | 0.05 | 0.096 | 0.871 |
| 2-div | - | - | - | - | 1.488 | 0.085 | 0.904 | 1.752 | 1.426 | 0.081 | 0.253 | 1.674 | 1.392 | 0.089 | 0.222 | 1.871 |
| 3-div | - | - | - | - | 2.473 | 0.094 | 1.53 | 2.752 | 2.396 | 0.105 | 0.407 | 2.669 | 2.351 | 0.122 | 0.352 | 2.871 |
| 1-SR | - | - | - | - | 6.32 | 0.479 | 3.424 | 7.994 | 6.351 | 0.47 | 3.665 | 8.051 | 6.512 | 0.473 | 3.737 | 8.147 |
| 2-SR | - | - | - | - | 5.999 | 0.465 | 3.215 | 7.637 | 6.029 | 0.459 | 3.326 | 7.709 | 6.187 | 0.461 | 3.505 | 7.798 |
| 3-SR | - | - | - | - | 5.944 | 0.46 | 2.939 | 7.536 | 5.969 | 0.453 | 2.978 | 7.623 | 6.123 | 0.455 | 3.158 | 7.705 |
| Crh | - | - | - | - | 2.088 | 0.143 | 1.548 | 3.408 | 2.181 | 0.177 | 1.815 | 8.51 | 2.223 | 0.187 | 1.373 | 10.442 |
| hROUGE | - | - | - | - | 0.069 | 0.014 | 0 | 0.138 | 0.071 | 0.015 | 0 | 0.153 | 0.071 | 0.014 | 0 | 0.134 |
| hBLEU | - | - | - | - | 0.008 | 0.01 | 0 | 0.07 | 0.008 | 0.01 | 0 | 0.077 | 0.01 | 0.01 | 0 | 0.099 |
| hBERTs | - | - | - | - | 0.382 | 0 | 0.361 | 0.396 | 0.376 | 0.001 | 0.342 | 0.399 | 0.38 | 0.003 | 0.352 | 0.398 |

| Statistic | NeuralChat | | | | Phi | | | | Qwen | | | | SOLAR | | | |
|---|---|---|---|---|---|---|---|---|---|---|---|---|---|---|---|---|
| | Mean | Std | Min | Max | Mean | Std | Min | Max | Mean | Std | Min | Max | Mean | Std | Min | Max |
| syC | 960.158 | 173.267 | 446 | 2794 | 677.627 | 123.055 | 192 | 2413 | 891.528 | 137.618 | 82 | 2984 | 613.768 | 94.711 | 32 | 1583 |
| lC | 584.639 | 97.224 | 350 | 1670 | 426.406 | 74.995 | 130 | 1658 | 555.051 | 80.995 | 62 | 1987 | 367.338 | 52.343 | 19 | 1077 |
| sC | 26.763 | 6.518 | 14 | 197 | 19.666 | 4.703 | 5 | 167 | 26.841 | 6.924 | 1 | 322 | 16.312 | 3.575 | 1 | 60 |
| FRE | 45.292 | 9.487 | -34.95 | 87.31 | 49.745 | 10.338 | -96.95 | 90.19 | 49.382 | 13.286 | -1279.49 | 112.59 | 42.051 | 10.153 | -5.68 | 90.6 |
| RC | 13.063 | 2.124 | 5 | 73 | 12.865 | 3.556 | 1 | 529 | 14.489 | 1.889 | 0 | 25 | | | | |
| POS-E | 4.446 | 0.054 | 3.63 | 4.959 | 4.468 | 0.069 | 1.491 | 5.046 | 4.453 | 0.066 | 1.282 | 5.822 | 4.444 | 0.244 | 0.059 | 4.882 |
| pPOS-E | 3.624 | 0.134 | 2.832 | 4.141 | 3.61 | 0.136 | 2.905 | 4.114 | 3.834 | 0.159 | 2.971 | 4.354 | 3.783 | 0.138 | 3.076 | 4.271 |
| Cr | 3.616 | 0.356 | 2.998 | 10.071 | 3.321 | 0.264 | 2.666 | 14.395 | 3.603 | 0.706 | 2.446 | 65.165 | 3.869 | 10.935 | 2.664 | 257.281 |
| dist | 3532.618 | 1773.924 | 1626 | 51518 | 3066.667 | 2052.043 | 1021 | 52597 | 3391.203 | 1815.441 | 1240 | 51930 | 3069.736 | 2632.549 | 1055 | 52550 |
| 1-div | 0.511 | 0.043 | 0.171 | 0.648 | 0.531 | 0.053 | 0.151 | 0.729 | 0.504 | 0.045 | 0.078 | 0.735 | 0.541 | 0.052 | 0.245 | 0.871 |
| 2-div | 1.401 | 0.077 | 0.448 | 1.614 | 1.435 | 0.081 | 0.393 | 1.721 | 1.391 | 0.084 | 0.187 | 1.705 | 1.451 | 0.075 | 0.902 | 1.871 |
| 3-div | 2.368 | 0.106 | 0.783 | 2.609 | 2.409 | 0.097 | 0.661 | 2.721 | 2.358 | 0.121 | 0.306 | 2.701 | 2.429 | 0.084 | 1.528 | 2.871 |
| 1-SR | 6.596 | 0.463 | 3.939 | 8.213 | 6.441 | 0.473 | 3.611 | 8.053 | 6.583 | 0.462 | 3.817 | 8.219 | 6.356 | 0.47 | 3.674 | 8.014 |
| 2-SR | 6.269 | 0.45 | 3.58 | 7.843 | 6.115 | 0.461 | 3.358 | 7.704 | 6.254 | 0.45 | 3.459 | 7.873 | 6.037 | 0.457 | 3.319 | 7.663 |
| 3-SR | 6.211 | 0.445 | 3.232 | 7.753 | 6.054 | 0.453 | 3.011 | 7.598 | 6.189 | 0.444 | 3.111 | 7.776 | 5.979 | 0.452 | 2.971 | 7.565 |
| Crh | 2.238 | 0.159 | 1.873 | 5.604 | 2.153 | 0.149 | 1.75 | 6.589 | 2.242 | 0.242 | 1.675 | 10.728 | 2.207 | 1.426 | 1.717 | 71.868 |
| hROUGE | 0.068 | 0.014 | 0 | 0.114 | 0.071 | 0.014 | 0 | 0.181 | 0.07 | 0.014 | 0 | 0.13 | 0.07 | 0.015 | 0 | 0.147 |
| hBLEU | 0.007 | 0.008 | 0 | 0.075 | 0.009 | 0.01 | 0 | 0.14 | 0.009 | 0.01 | 0 | 0.086 | 0.009 | 0.01 | 0 | 0.096 |
| hBERTs | 0.378 | 0.01 | 0.344 | 0.393 | 0.379 | 0.003 | 0.353 | 0.405 | 0.355 | 0.007 | 0.346 | 0.399 | 0.39 | 0.001 | 0.35 | 0.405 |

Table 7: Aggregated values of statistics from the human- and machine-generated texts in OpenTuringBench

Bench. It can be noticed a remarkable gap in realism, which renders data in OpenTuringBench closer to human-generated one, and thus more challenging to detect compared to other benchmarks akin to *TuringBench*, where "machines" can be easier perceived from less human-like and realistic generation patterns.

## E  Additional Insights on OTBDetector Evaluation

As OTBDetector learns similarity spaces, we additionally measured the **within-category compactness** and *across-category separation* of the learned spaces. The former is computed as the average pairwise similarity of the embeddings of objects sharing the same category:

$$intra(\mathcal{X}) = \frac{1}{|\mathcal{X}_k|} \sum_{y_k \in \mathcal{Y}} \sum_{X_i, X_j \in \mathcal{X}_k} sim(\mathbf{h}_i, \mathbf{h}_j), \quad (5)$$

whereas the latter is computed as the average pairwise similarity of the embeddings of objects belonging to two different categories:

$$inter(\mathcal{X}) = \frac{1}{|\mathcal{X}_h||\mathcal{X}_k|} \sum_{y_h, y_k \in \mathcal{Y}} \sum_{i \in h, j \in k} sim(\mathbf{h}_i, \mathbf{h}_j), \quad (6)$$

where $sim(\cdot, \cdot)$ denotes the cosine similarity function.

Here, we report the validation scores obtained by our best model settings, which has been used to perform all experiments in this work. For the Turing Test, we observe a within-category compactness of 0.960, indicating that texts within the same category (i.e., human or machine) are tightly grouped. Additionally, the separation between groups is pronounced, with a score of -0.824. Similarly, for the Authorship Attribution, we observe a compactness score of 0.975 and a separation of -0.128. These findings indicate that for both TT and AA, OTB-

> **Qualitative Comparison with Other Benchmarks**
>
> As the world continues to open its doors to travelers, the importance of being a responsible tourist cannot be overstated. From preserving the environment to respecting local cultures, the impact of tourism on destinations is undeniable. In this article, we explore the significance of responsible travel and offer tips for those seeking to make a positive difference during their journeys. Traveling responsibly means more than just visiting a destination and ticking off must-see attractions. It involves understanding the local culture, supporting local businesses, and minimizing the environmental footprint left behind. By embracing these principles, travelers can contribute to the preservation of natural resources and the well-being of communities they visit. [...]
>
> ---
>
> plan is to paint the ceiling, use a palette or brush add nice contrast floor. it with brush. spray paint. floor paint.fill that comes out will be color used for next step.you can apply if desired, but you want some depth floor, fill paint, floor.fill add

Figure 6: Machine-generated sample from *NeuralChat* in OpenTuringBench (top) and *gpt2_pytorch* in TuringBench (bottom).

Detector effectively learns to structure the semantic space for the downstream tasks, resulting in its strong detection and attribution capabilities. A visualization of the semantic spaces produced by OTBDetector for the AA and TT test sets is reported in Figures 7-8.

## F Details on Competing Detectors

- **Log-Likelihood** (Solaiman et al., 2019): This measure scores a text according to the average token-wise log probability yielded by a language model, with larger scores indicating a higher likelihood of the text being machine-generated.

- **Rank** (Gehrmann et al., 2019): This measure scores a text using the average rank value of its words computed, where individual ranks for each word are determined based on the preceding context. Smaller scores indicate a higher probability that the text is machine-generated.

- **Log-Rank** (Mitchell et al., 2023): Unlike the Rank, this variation first applies the log function to the individual rank of each word before averaging.

- **Entropy** (Gehrmann et al., 2019): Similarly to the rank score, this is obtained by averaging the entropy value of each word conditioned on the preceding context. MGT is likely to have a lower entropy score.

- **GLTR** (Gehrmann et al., 2019): This tool allows for getting the fraction of words that rank within a certain position (e.g., 10, 100, 1,000) in a given text, thus supporting the feature extraction for the subsequent classification tasks.

- **LRR** (Su et al., 2023): This score is a combination of the aforementioned Log-Likelihood and Log-Rank scores.

- **Fast-DetectGPT** (Bao et al., 2024): This approach leverages conditional probability curvature to determine word choice discrepancies between LLMs and humans, exploiting them to establish whether a text has been machine-generated.

- **OpenAI Detector** (Solaiman et al., 2019): This detector is a RoBERTa fine-tuning on data generated using the largest GPT2 model and is designed to predict whether a given text is machine-generated.

- **ChatGPT Detector** (Guo et al., 2023): This detector was developed by fine-tuning RoBERTa on the H3C (Human ChatGPT Comparison Corpus) dataset and is trained to distinguish between human- and ChatGPT-generated text.

- **LM Detector** (He et al., 2024): This is a fine-tuned DistilBERT with a classification module on top of it, optimized for distinguishing MGT.

- **DeTeCtive** (Guo et al., 2024): This is a recently developed end-to-end framework for AI-generated text detection that is based on multi-task auxiliary, multi-level contrastive loss to learn fine-grained features for distinguishing various writing styles and, hence, text generators.

| Statistic | Human | | | | Machines | | | |
|---|---|---|---|---|---|---|---|---|
| | Mean | Std | Min | Max | Mean | Std | Min | Max |
| syC | 745.494 | 484.927 | 4 | 2791 | 720.399 | 125.935 | 343.857 | 2230.143 |
| lC | 507.256 | 325.701 | 3 | 1814 | 445.344 | 73.780 | 185.714 | 1370.857 |
| sC | 27.084 | 18.37 | 1 | 151 | 20.867 | 5.829 | 4.857 | 160.857 |
| FRE | 63.676 | 11.613 | -68.26 | 104.64 | 47.401 | 14.257 | -339.216 | 91.377 |
| RC | 10.509 | 2.426 | 0 | 31 | 13.398 | 2.234 | 5 | 54.857 |
| POS-E | 4.479 | 0.089 | 3.914 | 4.968 | 4.456 | 0.084 | 3.099 | 4.854 |
| pPOS-E | 3.723 | 0.183 | 3.161 | 4.364 | 3.725 | 0.146 | 3.093 | 4.200 |
| Cr | 3.056 | 0.468 | 0.467 | 4.782 | 3.466 | 1.716 | 2.762 | 53.360 |
| dist | - | - | - | - | 3117.194 | 1988.217 | 1236.714 | 29061 |
| 1-div | - | - | - | - | 0.527 | 0.050 | 0.226 | 0.698 |
| 2-div | - | - | - | - | 1.426 | 0.081 | 0.649 | 1.668 |
| 3-div | - | - | - | - | 2.397 | 0.104 | 1.099 | 2.663 |
| 1-SR | - | - | - | - | 6.447 | 0.464 | 3.734 | 7.847 |
| 2-SR | - | - | - | - | 6.123 | 0.452 | 3.367 | 7.508 |
| 3-SR | - | - | - | - | 6.063 | 0.447 | 3.019 | 7.417 |
| Crh | - | - | - | - | 2.187 | 0.282 | 1.787 | 10.423 |
| hROUGE | - | - | - | - | 0.070 | 0.014 | 0 | 0.120 |
| hBLEU | - | - | - | - | 0.009 | 0.010 | 0 | 0.065 |
| hBERTs | - | - | - | - | 0.384 | 0.003 | 0.356 | 0.397 |

Table 8: Aggregated statistics from the human- and machine-generated texts in OpenTuringBench, main test set (i.e., E0). Values under column Machines are averages over the various LLMs' aggregated statistics.

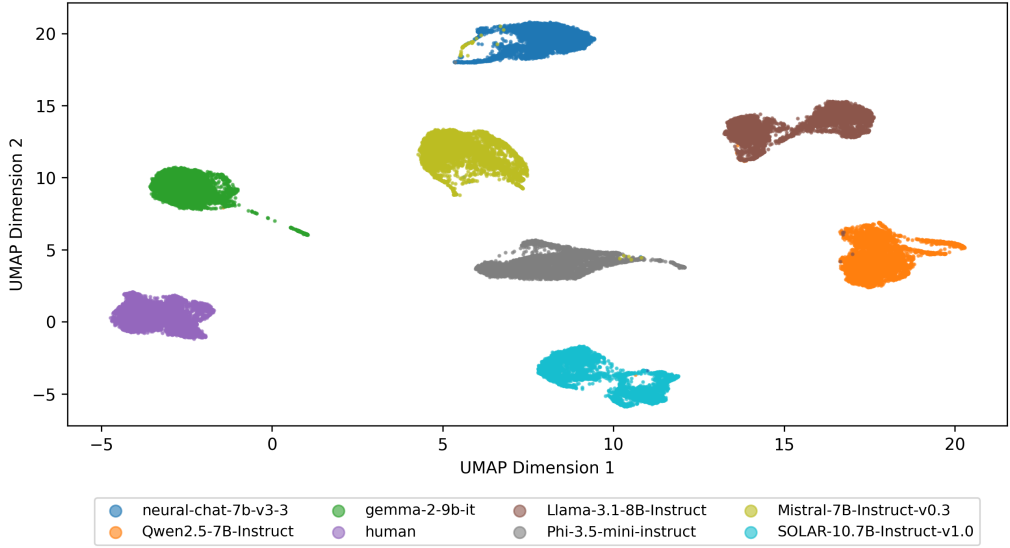

Figure 7: 2D UMAP visualization of the semantic space produced by OTBDetector for the AA test set.

| | Task | default | | | Higher Temp (0.7) | | | Higher Temp (1.0) | | | Larger Size | | | Self-Rewriting | | | Human Revision | | | Human Contin. | | |
|---|---|---|---|---|---|---|---|---|---|---|---|---|---|---|---|---|---|---|---|---|---|---|
| | | P | R | $F_1$ | P | R | $F_1$ | P | R | $F_1$ | P | R | $F_1$ | P | R | $F_1$ | P | R | $F_1$ | P | R | $F_1$ |
| Turing Test | Log-L | 0.973 | 0.989 | 0.981 | 0.970 | 0.905 | 0.936 | 0.925 | 0.342 | 0.500 | 0.973 | 0.989 | 0.981 | 0.971 | 0.917 | 0.943 | 0.916 | 0.303 | 0.455 | 0.943 | 0.461 | 0.619 |
| | Rank | 0.909 | 0.982 | 0.944 | 0.908 | 0.969 | 0.937 | 0.866 | 0.637 | 0.734 | 0.909 | 0.981 | 0.944 | 0.907 | 0.954 | 0.930 | 0.888 | **0.781** | 0.831 | 0.898 | 0.868 | 0.883 |
| | Log-R | 0.975 | 0.989 | 0.982 | 0.973 | 0.900 | 0.935 | 0.930 | 0.331 | 0.488 | 0.975 | 0.989 | 0.982 | 0.973 | 0.919 | 0.945 | 0.918 | 0.283 | 0.432 | 0.949 | 0.471 | 0.630 |
| | Entropy | 0.882 | **0.999** | 0.937 | 0.882 | 0.998 | 0.936 | 0.874 | 0.931 | 0.902 | 0.882 | **0.999** | 0.937 | 0.882 | **0.998** | 0.936 | 0.878 | 0.971 | **0.922** | 0.878 | **0.972** | **0.922** |
| | GLTR | 0.971 | 0.989 | 0.980 | 0.968 | 0.902 | 0.934 | 0.922 | 0.353 | 0.511 | 0.971 | 0.988 | 0.979 | 0.969 | 0.924 | 0.946 | 0.913 | 0.314 | 0.467 | 0.944 | 0.504 | 0.657 |
| | LRR | 0.972 | 0.982 | 0.977 | 0.969 | 0.888 | 0.927 | 0.922 | 0.333 | 0.490 | 0.972 | 0.981 | 0.977 | 0.970 | 0.919 | 0.944 | 0.911 | 0.289 | 0.439 | 0.950 | 0.536 | 0.685 |
| | FastDetect | 0.965 | 0.964 | 0.961 | 0.965 | 0.964 | 0.961 | 0.860 | 0.708 | 0.756 | 0.915 | 0.903 | 0.898 | 0.963 | 0.962 | 0.959 | 0.850 | 0.650 | 0.710 | 0.920 | 0.917 | 0.918 |
| | OAI-D | 0.994 | 0.767 | 0.866 | 0.989 | 0.462 | 0.630 | 0.961 | 0.123 | 0.218 | 0.993 | 0.731 | 0.842 | 0.992 | 0.591 | 0.741 | 0.959 | 0.117 | 0.208 | 0.938 | 0.075 | 0.139 |
| | GPT-D | 0.967 | 0.635 | 0.766 | 0.955 | 0.464 | 0.624 | 0.869 | 0.145 | 0.248 | 0.966 | 0.616 | 0.752 | 0.636 | 0.965 | 0.607 | 0.945 | 0.381 | 0.543 | 0.772 | 0.074 | 0.136 |
| | LM-D | **0.999** | **0.999** | **0.999** | **0.999** | 0.980 | **0.999** | **0.999** | **0.999** | **0.999** | **0.999** | **0.999** | **0.999** | **0.999** | 0.995 | **0.997** | **0.999** | 0.127 | 0.225 | **0.999** | 0.011 | 0.022 |
| | DeTeCtive | **0.999** | 0.979 | 0.989 | 0.998 | 0.993 | 0.996 | **0.999** | 0.968 | 0.984 | **0.999** | 0.998 | **0.999** | 0.998 | 0.732 | 0.844 | **0.999** | 0.139 | 0.244 | **0.999** | 0.127 | 0.225 |
| | **Ours** | **0.999** | **0.999** | **0.999** | **0.999** | **0.999** | **0.999** | 0.978 | 0.972 | 0.974 | **0.999** | **0.999** | **0.999** | 0.980 | 0.977 | 0.977 | 0.893 | 0.239 | 0.234 | 0.892 | 0.156 | 0.091 |

Table 9: ID and ID-V tasks (**E0**-**E4**) for Turing Test. Best scores are in bold, second-best underlined.

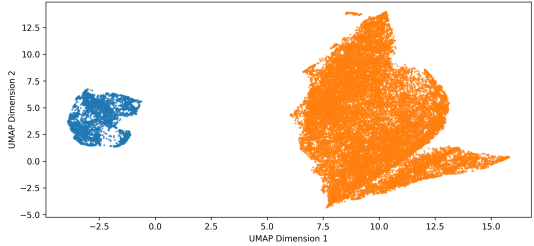

Figure 8: 2D UMAP visualization of the semantic space produced by OTBDetector for the Turing Test test-set. Blue points, resp. orange points, denote human-written texts, resp. machine-generated texts.

## G Additional Insights on Results

Tables 9-10 provide detailed results on Turing Test tasks following the same organization adopted for the Authorship Attribution tasks. This complements with our summary of detectors' performance on the Turing Test tasks presented in Sect. 5.1.

| | Test Task | Out-of-Domain Text | | | Unseen Model | | |
|---|---|---|---|---|---|---|---|
| | Detector | P | R | $F_1$ | P | R | $F_1$ |
| Turing Test | Log-L | 0.999 | 0.985 | 0.993 | 0.837 | 0.999 | 0.911 |
| | Rank | 0.999 | 0.991 | 0.996 | 0.592 | 0.999 | 0.744 |
| | Log-R | 0.999 | 0.987 | 0.993 | 0.850 | 0.999 | 0.919 |
| | Entropy | 0.999 | 0.999 | **0.999** | 0.516 | 0.999 | 0.680 |
| | GLTR | 0.999 | 0.984 | 0.992 | 0.827 | 0.998 | 0.905 |
| | LRR | 0.999 | 0.985 | 0.992 | 0.836 | 0.999 | 0.910 |
| | FastDetect | 0.999 | 0.999 | **0.999** | 0.887 | 0.854 | 0.850 |
| | OAI-D | 0.999 | 0.568 | 0.724 | 0.960 | 0.825 | 0.887 |
| | GPT-D | 0.999 | 0.576 | 0.731 | 0.804 | 0.628 | 0.705 |
| | LM-D | 0.999 | 0.976 | 0.988 | 0.999 | 0.999 | **0.999** |
| | DeTeCtive | 0.999 | 0.898 | 0.946 | 0.998 | 0.999 | 0.998 |
| | **Ours** | 0.999 | 0.963 | 0.981 | 0.999 | 0.999 | **0.999** |

Table 10: OOD tasks (**E5**-**E6**) for Turing Test. Best $F_1$ scores are in bold, second-best underlined.

16